\documentclass[10pt,twocolumn,letterpaper]{article}
\usepackage[pagenumbers]{cvpr}
\usepackage{graphicx,amsmath,amssymb,booktabs,array,tikz,pgfplots,capt-of}
\usetikzlibrary{positioning,arrows.meta,fit,backgrounds,calc}
\pgfplotsset{compat=1.16}
\usepackage[utf8]{inputenc}\usepackage[T1]{fontenc}
\definecolor{cvprblue}{rgb}{0.21,0.49,0.74}
\definecolor{familiar}{HTML}{2A78D6}\definecolor{heldout}{HTML}{EB6834}
\usepackage[breaklinks,colorlinks,allcolors=cvprblue]{hyperref}
\def\confName{CVPR}\def\confYear{2027}
\newcommand{\para}[1]{\smallskip\noindent\textbf{#1}\hspace{0.4em}}

\title{Four Logged Interactions for Control of New Shapes:\\A Frozen Vision-Language Model Meets Latent World Models}
\author{Shicheng Yin$^{1*}$\quad Yueyang Ou$^{2*}$\quad Chen Ming Chun$^{1*}$\quad Kaixuan Yin$^{1}$\quad Yang Liu$^{1,3\dagger}$\\
$^{1}$Sun Yat-sen University\quad $^{2}$Tianjin University\quad $^{3}$X-Era AI Lab\\
{\small \{yinshch5, chenmj88, yinkx3\}@mail2.sysu.edu.cn,\ \ liuy856@mail.sysu.edu.cn,\ \ ouyueyang@tju.edu.cn}\\
{\small Code: \href{https://github.com/HCPLab-SYSU/PREP}{\textcolor{black}{github.com/HCPLab-SYSU/PREP}}}}

\begin{document}
\twocolumn[{\renewcommand\twocolumn[1][]{#1}\maketitle
\begin{center}
\includegraphics[width=.9\textwidth]{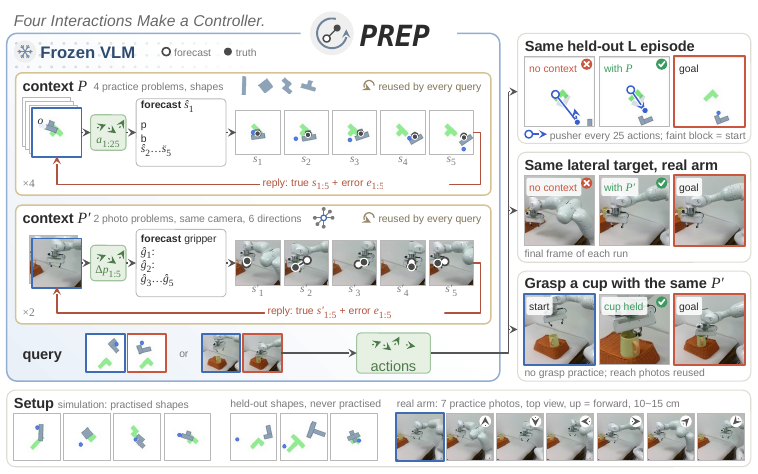}
\captionof{figure}{\textbf{PREP at work, with actual model outputs.} \emph{Left, top:} one of the four simulation practice exchanges that form the record $P$ (labeled \emph{context} in the figure): from the frames and the actions $a_{1:25}$, the frozen VLM forecasts the pusher and block at five future times (hollow markers) and is then shown the true states (solid markers) and its errors. \emph{Left, bottom:} a control query appends the current images and a goal image to the same record, and the model returns actions directly. \emph{Right:} the same held-out L episode without and with $P$; the same lateral reaching target on the real arm without and with $P'$; and a cup grasp with the same $P'$.}
\label{fig:teaser}
\end{center}}]

{\let\thefootnote\relax\footnotetext{$^{*}$Equal contribution.\quad $^{\dagger}$Corresponding author.}}
\begin{abstract}
We study how a frozen vision-language model uses a small record of physical experience to control objects beyond those in the record. PREP turns four logged action sequences into forecasting exchanges with outcome feedback, freezes the transcript, and reuses it for direct goal-image control. On 350 PushObj episodes, adding this record to Claude Opus~5.5 raises success from 12.9\% to 65.4\%, with unchanged weights and control queries. On three shapes excluded from task-specific data, success reaches 62.7\%, versus 14.7--28.7\% for DINO-WM, LeWM, and JEPA-WM trained on 20{,}000 trajectories. The strongest world model remains competitive on training shapes: 63.0\% versus PREP's 67.5\%. Thus useful control already emerges from logged transitions, with the full practice record performing best. A seven-photograph instantiation on a Franka arm achieves 10/20 successes versus 7/20 without the record and 11/20 for V-JEPA~2-AC; identical-input replay localizes the benefit to lateral action commands.
\end{abstract}

\section{Introduction}
\label{sec:intro}

A frozen vision-language model (VLM) brings broad visual and spatial knowledge to a manipulation scene, yet when shown current frames and a goal image and asked for numerical actions, it makes a poor controller. Without experience on that specific system, it does not know how action values map to displacement, how image and control axes align, how far the actuator moves per step, or how a contacted object responds.

\begin{figure*}[t]\centering
\includegraphics[width=\textwidth]{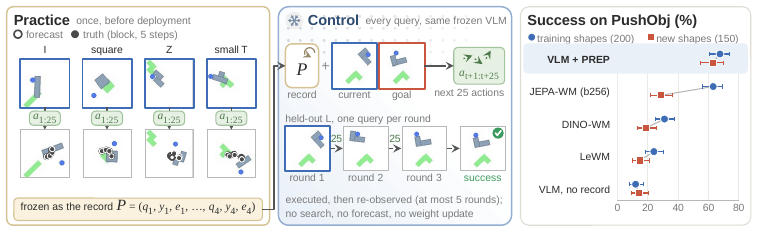}
\caption{\textbf{PREP in outline, and the main result.} \emph{Left:} before deployment, a frozen VLM answers four forecasting problems cut from logged trajectories and is shown the true outcomes and its errors; the conversation is frozen as the record $P$. \emph{Right:} success on 350 PushObj episodes with Wilson 95\% intervals. On three shapes excluded from task-specific data, success reaches 62.7\%, versus 14.7--28.7\% for DINO-WM, LeWM, and JEPA-WM trained on 20{,}000 trajectories.}
\label{fig:overview}
\end{figure*}

We ask whether a frozen VLM can acquire the local physical calibration it lacks from a handful of logged interactions placed in its context, and how far it then reaches. We study in-context physical practice (\textbf{PREP}; \cref{fig:teaser,fig:overview}), which builds a fixed conversational prefix before deployment from short segments of logged trajectories (\cref{sec:method}). Every control query appends the latest observations and a goal image to this record, and the VLM writes the next actions directly, with no planner, no candidate scoring, no test-time forecast, and no weight update; nothing from deployment is ever added to the record.

\begin{figure}[t]\centering
\includegraphics[width=\columnwidth]{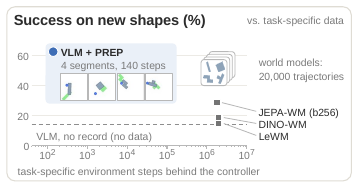}
\caption{Task-specific data and success on the new shapes. The record uses four logged segments of 35 actions; each world model trains on 20{,}000 trajectories of 100 steps (JEPA-WM: 18{,}000 plus 2{,}000 for validation). Pretraining data are not counted for any model.}
\label{fig:data}
\end{figure}

We test PREP on PushObj, a shape-varying Push-T benchmark~\cite{chi2023,maes2026b} with 350 episodes across four training shapes and three new shapes absent from all task-specific data (L, T, +; \cref{fig:shapes}). Queried without a record, Claude Opus~5.5 succeeds in 12.9\% of episodes, below every world model we train on the task. Prepending a PREP record built from $K{=}4$ logged segments, one per training shape, raises the same model to 65.4\% (67.5\% on training and 62.7\% on new shapes; \cref{tab:main}). Of LeWM, DINO-WM, and JEPA-WM, trained on 20{,}000 trajectories of the training shapes and planned with CEM under the same observation schedule (\cref{sec:setup,tab:interface}), the strongest, JEPA-WM at batch size 256, is level with PREP on the training shapes (63.0\%), but on the new shapes the three reach 14.7--28.7\% (\cref{fig:overview,fig:data}).
Local transition examples already enable substantial control, while the forecast-and-feedback construction gives the strongest observed record.

\para{Contributions.} (1) PREP, a weight-free procedure that turns a frozen VLM into a direct goal-image controller from four logged interactions. (2) A controlled 350-episode comparison with three latent world models trained on the same shapes, including new shapes, diagnosing where the trained route loses them. (3) Matched prefix studies separating the record's content from its length. (4) A real-arm instantiation from seven photographs, evaluated in executed trials and replay.

\section{Related Work}
\label{sec:related}

\para{Latent world models.}
Model-based control learns a dynamics model from interaction data and plans through it~\cite{chua2018,hafner2019,hansen2024,hafner2020dreamer,hafner2025dreamerv3}. Joint-embedding predictive architectures~\cite{lecun2022,assran2023ijepa,bardes2024vjepa} predict in latent space rather than pixels: DINO-WM~\cite{zhou2025} trains an action-conditioned predictor on frozen DINOv2 features~\cite{oquab2024}, JEPA-WM~\cite{terver2026} studies which design choices matter for physical planning, LeWM~\cite{maes2026a} trains encoder and predictor jointly from pixels, and V-JEPA~2-AC~\cite{assran2025} trains an action-conditioned predictor on robot video. These methods acquire local dynamics through training and act through search; we use all four as reference controllers with their own planning procedures.

\para{VLMs for robotic control.}
Vision-language-action models such as RT-2~\cite{brohan2023} and OpenVLA~\cite{kim2024} fine-tune pretrained backbones on robot demonstrations. Without weight updates, frozen models are typically paired with external structure: generating policy code~\cite{liang2023}, specifying 3D value maps or keypoint constraints~\cite{huang2023,huang2024}, iteratively refining visual action proposals in PIVOT~\cite{nasiriany2024}, or scoring candidate actions rolled out by a video prediction model in VLMPC~\cite{zhao2024}. PREP writes numerical actions directly from the frozen VLM, adding no code synthesis, value map, external predictor, or action head; only the context changes.

\para{In-context learning and feedback.}
Pretrained sequence models adapt from examples in context~\cite{brown2020,mirchandani2023,benechehab2025dicl,liu2024dynamical}. In manipulation, in-context learning enables imitation from demonstrations~\cite{dipalo2024,yin2025roboprompt} and policy improvement from previous attempts~\cite{vandermerwe2025}, while language agents revise outputs from feedback~\cite{madaan2023,shinn2023} or update a textual world model from prediction errors during deployment in text environments~\cite{zhang2026}. PREP instead forms its record before deployment from logged transitions, with no demonstration of the control task and no attempt at it, and keeps it fixed across goals, episodes, and new objects. In system-identification terms~\cite{ljung1999}, the practice problems probe a local input-output response, and we measure the resulting controller.

\section{PREP: In-Context Physical Practice}
\label{sec:method}

\subsection{The practice record}

Before deployment, PREP builds a \emph{practice record} $P$ from $K$ segments cut from logged trajectories (Algorithm~1). On PushObj (\cref{sec:pushobj}), each segment spans 35 environment steps, grouped by five into \emph{merged steps}. For each segment $i=1,\dots,K$, the practice query $q_i$ gives three frames spaced one merged step apart, the ten past actions executed between them, and the next 25 actions $a_{1:25}$ (five merged steps), and asks the frozen VLM (Claude Opus~5.5) to forecast the resulting pusher and block states at the five future merged steps ($y_i$). Since the actions are prescribed, the model chooses nothing and pursues no goal. The query $q_i$ omits simulator coordinates and pusher velocity, so the model must read positions from the frames, while measured states are provided only in the feedback after it commits to $y_i$.

The feedback turn $e_i$ then supplies the true states at those five steps, the per-step numerical errors of $y_i$, and the true future frames (\cref{fig:teaser}, top left). Each exchange is appended so subsequent problems are answered with earlier forecasts and their corrections in context; after the $K$ exchanges,
\begin{equation}
P=(q_1,y_1,e_1,\;\dots,\;q_K,y_K,e_K)
\end{equation}
is frozen as the prefix for control. On PushObj, the \emph{standard record} uses $K=4$ segments, one per training shape (I, square, Z, and small T) and none from the new shapes (\cref{fig:overview}), practiced once at maximal reasoning effort with no goal image, reward, or demonstration.

\begin{table}[t]
\small
\begin{tabular}{@{}p{\columnwidth}@{}}
\toprule
\textbf{Algorithm 1}\ \ PREP\\
\midrule
\textbf{Practice} (once, from $K$ logged segments)\\
\ 1:\ \ $P\leftarrow()$\\
\ 2:\ \ \textbf{for} $i=1,\dots,K$ \textbf{do}\\
\ 3:\ \ \quad $q_i\leftarrow$ frames, past actions, and next actions of segment $i$\\
\ 4:\ \ \quad $y_i\leftarrow\mathrm{VLM}(P\,\Vert\,q_i)$\hfill forecast of future states\\
\ 5:\ \ \quad $e_i\leftarrow$ true states, per-step errors of $y_i$, true frames\\
\ 6:\ \ \quad $P\leftarrow P\,\Vert\,(q_i,y_i,e_i)$\\
\ 7:\ \ freeze $P$\\[2pt]
\textbf{Control} (every episode, goal image $o_g$)\\
\ 8:\ \ \textbf{repeat} at most five times, until the goal is reached\\
\ 9:\ \ \quad $c_t\leftarrow$ three latest frames, actions between them, $o_g$\\
10:\ \ \quad $a_{t+1:t+25}\leftarrow\mathrm{VLM}(P\,\Vert\,c_t)$\hfill $P$ is never modified\\
11:\ \ \quad execute the 25 actions\\
\bottomrule
\end{tabular}
\end{table}

\subsection{Control queries}

At control time, each decision independently continues the frozen record $P$ as its prefix. At environment step $t$, query $c_t=(o_{t-10},o_{t-5},o_t,a_{t-9:t},o_g)$ shows the three latest frames, the ten actions executed between them, and a goal image $o_g$, asking for the next 25 actions. As in practice, no pusher position or velocity or environment rules are given, matching LeWM's inputs.

Queried at high reasoning effort, the model writes directly,
\begin{equation}
a_{t+1:t+25}=\mathrm{VLM}\big(P\,\Vert\,c_t\big),
\end{equation}
with no search, no candidate actions proposed, no forecast, and no weight update. The 25 actions are executed; if the goal is not reached, $c_{t+25}$ is posed against the same $P$, for at most five rounds per episode. Prior control exchanges are not kept in context, so $P$ stays fixed across decisions, episodes, goals, and shapes. Without PREP, the \emph{no-record} controller omits only $P$, leaving the model, query wording, observations, and settings unchanged; every other VLM condition below likewise varies only the prefix.

\subsection{The robot record}
\label{sec:robotrecord}

On the Franka arm (\cref{sec:robot}), the record $P'$ is built from seven photographs taken in the test scene and camera pose while moving only the end effector (start pose; 15\,cm forward, backward, left, right; 10\,cm forward-right, backward-left). They form $K{=}2$ problems, each ordering six photographs into a sequence where consecutive measured end-effector state differences define five actions. In each, the model receives the initial photograph, its end-effector state, and the five actions, and forecasts in one answer at high reasoning effort the end-effector position and eight image keypoints after each action, which a separate Claude Opus~5.5 call marks consistently on the gripper, wrist, and table objects across a problem's photographs without segmentation. Feedback then shows the true positions and keypoints, its errors, and the five true photographs with predicted and true points drawn and joined by lines. After two exchanges, $P'$ is frozen (about 14K tokens).

Control queries append the current image, end-effector state, and goal image to $P'$, and the model writes one end-effector displacement and gripper command.

\section{Experimental Setup}
\label{sec:setup}

\subsection{PushObj}
\label{sec:pushobj}

\begin{figure}[t]\centering
\includegraphics[width=\columnwidth]{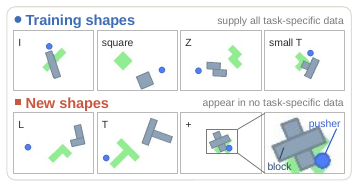}
\caption{PushObj start frames. The four training shapes supply the world models' 20{,}000 trajectories and the four practice problems; the three new shapes appear in no task-specific data. The pusher is the small blue disk.}
\label{fig:shapes}
\end{figure}

PushObj is a shape-varying version of Push-T~\cite{chi2023} in the stable-worldmodel platform~\cite{maes2026b}, observed through $224{\times}224$ images (\cref{fig:shapes}). Four \emph{training shapes} (I, square, Z, and small T) supply all task-specific data: the world models' 20{,}000 logged trajectories (5{,}000 of 100 steps per shape) and the four practice problems in $P$. Neither the world models nor $P$ ever see the three \emph{new shapes} (L, T, and +): T is a larger relative of the practiced small T, whereas L and + have no training-shape counterpart. Because Push-T is public, it may appear in VLM pretraining data; whatever the model retains from it, without the record it succeeds in 12.9\% of PushObj episodes.

Test problems are 25-step segments from separately collected test trajectories, each with a start and a goal. We discard segments where the pusher leaves the workspace, the block is pinned against a wall, or the pusher must first circle to the far side of the block, and balance the difficulty distribution across shapes, keeping 50 problems per shape. Across the 350 problems, mean block displacement is 45.2 pixels and mean rotation is 37.4$^\circ$.

Evaluation runs on the same 350 episodes for every controller, from each problem's start toward its goal, executing 25 actions between observations for at most five rounds. Under the official Push-T criterion, episodes succeed when the Euclidean distance between current and goal pusher and block positions is below 20 (world units, in a $512{\times}512$ workspace) and the block angle lies within 20$^\circ$ of the goal, modulo the shape's rotational symmetry.

\subsection{World-model reference controllers}
\label{sec:wm}

We compare PREP with three latent world models: DINO-WM~\cite{zhou2025}, LeWM~\cite{maes2026a}, and JEPA-WM~\cite{terver2026} (\cref{tab:interface}). All three train on the 20{,}000 trajectories of the training shapes. DINO-WM trains a predictor on frozen, pretrained DINOv2~\cite{oquab2024} features for 20 epochs. LeWM trains its encoder and predictor jointly from pixels, without pretraining, for 62 epochs. JEPA-WM, run with its official code and Push-T optimization settings, also trains a predictor on frozen DINOv2. For JEPA-WM we set aside 2{,}000 trajectories, drawn per shape, for validation, train at total batch size 256, and use the lowest-validation-loss checkpoint. No checkpoint is selected using the new shapes.

The comparison keeps deployment observations explicit (\cref{tab:interface}). DINO-WM and JEPA-WM receive three images, pusher position and velocity, and actions; LeWM and the VLM use images and actions without pusher state. During preparation, PREP additionally receives measured block and pusher states in feedback, whereas the world models train without block-pose labels. The shared task and observation cadence therefore coexist with different representations, preparation supervision, and pretraining.

\begin{table}[t]
\centering\footnotesize\setlength{\tabcolsep}{3pt}
\begin{tabular}{@{}llll@{}}
\toprule
Controller & Task-specific data & Input per decision & Actions from\\
\midrule
LeWM & 20{,}000 trajectories & 3 frames, goal & CEM search\\
DINO-WM & 20{,}000 trajectories & + pusher state & CEM search\\
JEPA-WM & 18{,}000 (+2{,}000 val.) & + pusher state & CEM search\\
VLM, no record & none & 3 frames, goal & direct output\\
VLM + PREP & 4 logged segments & 3 frames, goal & direct output\\
\bottomrule
\end{tabular}
\caption{What each PushObj controller uses. All receive the actions executed between the three frames. Every controller observes once per 25 actions, at most five times per episode.}
\label{tab:interface}
\end{table}

The world models search over five merged steps (25 environment actions) with the cross-entropy method (CEM)~\cite{rubinstein1999cross,deboer2005tutorial}. Each iteration samples 300 action sequences and retains the best 30, for 30 iterations, the same budget for every model. The cost is the distance between predicted final features and goal-image features; DINO-WM and JEPA-WM add an equally weighted pusher position and velocity term. The 25 planned actions run before replanning, at most five times per episode, so every controller sees the scene equally often. JEPA-WM ran on a different GPU set with TF32 matrix multiplication, the others in fp32. As a check, with a single plan on the 50 L episodes, its batch-256 checkpoint succeeds 11 times in fp32 and 10 in TF32. Because shapes vary and 25 actions run between observations, absolute success rates should not be compared with published Push-T numbers.

\para{Statistics and release.} We report successes with their denominators, Wilson 95\% intervals, and two-sided Fisher exact tests. The following are released in an anonymous repository: the verbatim practice records and control prefixes for every condition, all practice and test problems, the control and evaluation scripts, and every model output and success judgment.

\section{Results}
\label{sec:results}

\subsection{Prediction and control from the same record}
\label{sec:forecast}

We first examine the record in the setting where it is constructed: predicting consequences of supplied actions. The probe contains five problems per shape, 35 in all, drawn from the 350 test problems. From the standard record, the model predicts pusher and block poses at five merged-step boundaries, at high reasoning effort. Each problem receives three independent answers; each answer is scored separately and the resulting errors are averaged.

To compare placements in a learned visual space, we render each predicted pose and encode the image using the reference world model's encoder. The reference predicts features directly, and both outputs are compared with the encoded true frame. DINOv2 supplies 256 patches of 384 dimensions; LeWM supplies one 192-dimensional projection vector. Rendering supplies the known object geometry to the VLM route. Thus this probe measures how well its numerical placement reconstructs a future scene, rather than asserting equal native image-generation interfaces.

\begin{table}[t]
\centering\small\setlength{\tabcolsep}{3pt}
\begin{tabular}{@{}llcc@{}}
\toprule
Space & Forecast & Training (20) & New (15)\\
\midrule
DINOv2 & DINO-WM & 0.440 & 0.611\\
 & VLM pose-rendered & 0.118 & 0.312\\
LeWM & LeWM & 0.269 & 0.425\\
 & VLM pose-rendered & 0.079 & 0.143\\
\bottomrule
\end{tabular}
\caption{Fifth-step feature MSE of predicted scenes. VLM pose-rendered forecasts have lower error on 19/20 training problems in both spaces, and on 13/15 new problems in DINOv2 and 12/15 in LeWM. Errors are averaged over three independently scored VLM answers per problem.}
\label{tab:forecast}
\end{table}

The practiced model gives the closer scene on 32/35 problems in DINOv2 and 31/35 in LeWM (\cref{tab:forecast}), including a majority of new-shape problems in each space. In DINOv2, its feature MSE over five steps is 0.053, 0.069, 0.126, 0.121, and 0.201, versus DINO-WM's 0.190, 0.284, 0.332, 0.398, and 0.513. Both accumulate error, but the pose-rendered predictions remain closer throughout the rollout.

Preparation effort also changes subsequent prediction. On these 35 problems, with high-effort testing fixed, maximal rather than high-effort preparation reduces fifth-step keypoint error from 9.70 to 2.41 pixels, improving 34/35 problems. When the model writes poses instead, the corresponding error is 13.76 versus 4.75 pixels, with maximal-effort preparation better on 31/35.

\subsection{Does the record make a controller that reaches new shapes?}
\label{sec:main}

\begin{table}[t]
\centering\small\setlength{\tabcolsep}{3.5pt}
\begin{tabular}{@{}lccc@{}}
\toprule
Controller & Training & New & All\\
 & (200) & (150) & (350)\\
\midrule
VLM, no record & 12.0 {\scriptsize(24)} & 14.0 {\scriptsize(21)} & 12.9 {\scriptsize(45)}\\
\textbf{VLM + PREP record} & \textbf{67.5} {\scriptsize(135)} & \textbf{62.7} {\scriptsize(94)} & \textbf{65.4} {\scriptsize(229)}\\
\midrule
LeWM + CEM & 24.0 {\scriptsize(48)} & 14.7 {\scriptsize(22)} & 20.0 {\scriptsize(70)}\\
DINO-WM + CEM & 31.0 {\scriptsize(62)} & 18.7 {\scriptsize(28)} & 25.7 {\scriptsize(90)}\\
JEPA-WM + CEM, batch 256 & 63.0 {\scriptsize(126)} & 28.7 {\scriptsize(43)} & 48.3 {\scriptsize(169)}\\
\bottomrule
\end{tabular}
\caption{Success (\%) within five rounds on the 350 PushObj episodes, with success counts in parentheses. The two VLM rows differ only in the practice record. Training shapes are I, square, Z, and small T; new shapes are L, T, and +.}
\label{tab:main}
\end{table}

\Cref{tab:main} and \cref{fig:overview} (right) give the main comparison. Without the record, the frozen VLM succeeds in 45 of 350 episodes (12.9\%): 12.0\% on the training shapes and 14.0\% on the new shapes. Adding the frozen four-problem record $P$ raises success to 229 of 350 (65.4\%, $p<10^{-40}$): 67.5\% on the training shapes and 62.7\% on the new shapes. \Cref{fig:teaser} (top right) shows one L episode that fails without the record and succeeds with it.

On the training shapes (I, square, Z, and small T), the world models trained on them reach 24.0\% to 63.0\%. The JEPA-WM batch-256 run provides the strongest reference. Its training-shape success is close to PREP's (63.0\% versus 67.5\%, $p=0.40$), showing that this controller can solve the task at a comparable rate on the objects supplying its experience.

\begin{table}[t]
\centering\small\setlength{\tabcolsep}{3.2pt}
\begin{tabular}{@{}lccccccc@{}}
\toprule
JEPA-WM & I & Sq. & Z & sm.\,T & L & T & +\\
\midrule
batch 256 & 30 & 27 & 35 & 34 & 15 & 14 & 14\\
\bottomrule
\end{tabular}
\caption{JEPA-WM successes per shape, out of 50 episodes each.}
\label{tab:jepashape}
\end{table}

On the new shapes (L, T, and +), the two routes diverge. The record shows only training shapes and is never extended during evaluation. PREP's gain there therefore comes from four practice problems on other shapes, combined with the model's pretrained knowledge of shapes. JEPA-WM at batch 256 reaches 28.7\% against 62.7\% ($p<10^{-8}$; Wilson intervals 22.0--36.4\% and 54.7--70.0\%). DINO-WM and JEPA-WM both encode images with frozen, pretrained DINOv2, so pretrained visual features alone do not prevent the drop.
The drop is not tied to one shape. On the first 20 test episodes of each new shape (\cref{fig:matched}), the record succeeds on 14 T, 14 +, and 10 L episodes, against 5, 4, and 2 without it. T is a larger relative of the practiced small T, but + and L have no practiced counterpart, and the record adds 10 and 8 successes there.
\subsection{Is it the content of the record?}
\label{sec:prefix}

\begin{figure}[t]\centering
\includegraphics[width=\columnwidth]{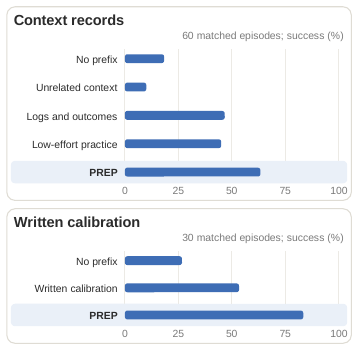}
\caption{\textbf{Quantitative prefix comparisons.} Top: success rates of no prefix, unrelated context, logs and outcomes, low-effort practice, and PREP on the same 60 new-shape episodes; PREP reaches 63.3\% versus 18.3\% without a prefix. Bottom: success rates of no prefix, written calibration, and PREP on a separate matched 30-episode set. Local transition examples already enable substantial control, while the forecast-and-feedback construction gives the strongest observed record.}
\label{fig:matched}
\end{figure}

Success rates for all prefix conditions are shown in \cref{fig:matched}. All five conditions use the same frozen model, control query, high reasoning effort, and 125-action budget. The no-prefix and standard-record outcomes are the corresponding episodes from the main evaluation; the additional conditions change only the prefix.

\para{Physical content.} An unrelated conversation matched to the standard record's length gives 6/60 successes, compared with 11/60 without a prefix ($p=0.30$) and 38/60 with PREP ($p<10^{-8}$). The standard record also exceeds no prefix on this subset ($p<10^{-6}$), with gains on each shape. Context length alone therefore fails to reproduce the benefit of physical experience.

\para{Observed outcomes versus prediction practice.} The logs-only prefix contains the same four problems (I02, square02, Z02, small\_tee02), their images and action sequences, and the measured outcomes and true frames. The model reads and acknowledges them without making forecasts or receiving forecast errors. It succeeds on 11 T, 10 +, and 7 L episodes: 28/60, compared with PREP's 14, 14, and 10, or 38/60. Reading local transitions already produces substantial control. PREP's additional ten successes favor the complete construction, though the aggregate Fisher comparison remains suggestive ($p=0.10$).

\para{Preparation effort.} Low-effort practice uses the same four problems and feedback, produces brief forecasts, and still uses high effort for control. It succeeds 9 times on each new shape, 27/60 in all, compared with 38 for standard practice ($p=0.07$). On a separate 30-episode small-T/T set, standard, low-effort, and no-prefix control yield 22, 11, and 6 successes; standard versus low has $p=0.009$. Together these comparisons favor standard preparation.

\para{Written calibration.}
We replace $P$ with a written calibration of quantities that can be read off the practice feedback: how action values convert to pixel displacement, the farthest the pusher moves in one merged step, the directions of the image axes, the pusher and block sizes, and that a contacted block roughly follows the pusher; the prefix contains only this text. On ten episodes each of I, square, and T (\cref{fig:matched}), calibration doubles success over no prefix (16 versus 8 of 30; $p=0.06$) and comes within one and two successes of the record on I and square. On T it succeeds 3 times in 10, against 9 with the record ($p=0.02$); over all 30 the record leads 25 to 16 ($p=0.025$). Written constants thus recover a substantial part of the effect. With ten episodes per shape we draw no shape-specific conclusions, but the record was ahead on the one new shape in this set.

\subsection{Why do the world models lose new shapes?}
\label{sec:probe}

\begin{figure*}[t]\centering
\includegraphics[width=.6\textwidth]{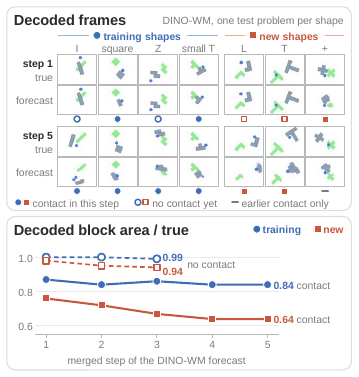}
\caption{DINO-WM forecasts decoded to images on all 350 problems: the area of the predicted block relative to the decoding of the true frame, split by whether the pusher touches the block in that step. Top: the first test problem of each shape in which the pusher touches the block, at merged steps 1 and 5; each pair shows the decoding of the true frame above the decoded forecast, and the marker below gives that step's contact state.}
\label{fig:probe}
\end{figure*}

DINO-WM has an image decoder, so we can inspect its predicted scenes. On all 350 problems we roll DINO-WM open loop for five merged steps from one frame and 25 actions, decode the predicted features, and count pixels within RGB distance 40 of the block color, relative to the same count in the decoded encoding of the true frame, splitting steps by whether the pusher touches the block (\cref{fig:probe}).

While the block is untouched, the forecast keeps new shapes almost intact: relative area is 0.99--1.00 for training and 0.94--0.98 for new shapes. The gap opens under contact: training blocks retain 0.84--0.87, whereas new blocks drop from 0.76 at step~1 to 0.64 by step~5. The forecast thus keeps an unfamiliar block while it is untouched and loses part of it once pushed. This points to predicted interaction, not the encoding of a new shape, as where the model fails, although the decoder's response to unusual predicted features may contribute. Because the same decoder renders untouched new blocks nearly whole, a decoder unable to draw new shapes does not by itself explain the drop. Contact forecasts are exactly those a planner relies on to rank candidate actions. JEPA-WM and LeWM have no decoder, so we do not inspect their forecasts this way; in control they show the same drop.

\subsection{Does it work on a real arm?}
\label{sec:robot}

We follow the real-robot protocol of V-JEPA~2~\cite[Sec.~4]{assran2025} on a Franka Emika Panda with its standard gripper, in a laboratory not included in DROID~\cite{khazatsky2024}. An uncalibrated monocular RGB camera views the arm from the front right; for both controllers, images and goal images are mirrored horizontally and the lateral components of state and action negated, so that the view resembles DROID's left camera. Reaching uses a locked gripper, four start poses, and targets 15--20\,cm away in one of six directions (forward, backward, left, right, up, down); all four steps are executed, and success means ending within 4\,cm of the target. Cup grasping varies the cup position and start pose; after four open-gripper steps the gripper closes, and success means the cup is held. Each step executes one action, an end-effector displacement capped at 7.5\,cm per axis with fixed orientation plus a gripper command, before the next decision. Every goal is given as an image.

The VLM uses no robot data beyond the seven practice photographs behind $P'$ (\cref{sec:robotrecord}), no task-specific training, and no reward. Each query appends the current image, the goal image, and the end-effector state to $P'$; the model reads object and gripper positions from the image itself and writes the action directly, with no search, answering each step from the fixed record alone at high reasoning effort. V-JEPA~2-AC~\cite{assran2025} uses its public weights, reproduced on the same arm and camera. Its frozen ViT-g video encoder has over one billion parameters and its action-conditioned predictor about 300 million; trained on less than 62 hours of unlabeled DROID video, it plans with horizon-one CEM (800 samples, ten iterations, top ten), minimizing the L1 distance to the goal representation. Both controllers receive the end-effector state.

\begin{table}[t]
\centering\small\setlength{\tabcolsep}{5pt}
\begin{tabular}{@{}lccc@{}}
\toprule
Controller & Reaching & Cup grasping & Total\\
\midrule
VLM, no record & 5/10 & 2/10 & \phantom{0}7/20\\
\textbf{VLM + PREP record} & 7/10 & 3/10 & 10/20\\
V-JEPA~2-AC & 7/10 & 4/10 & 11/20\\
\bottomrule
\end{tabular}
\caption{Real-arm trials, ten per task per controller.}
\label{tab:robot}
\end{table}

\para{Executed trials.} In \cref{tab:robot}, practice adds two reaching successes and one grasp, and the practiced VLM matches V-JEPA~2-AC in reaching with one fewer grasp. With ten trials per task these totals are indicative only (Fisher $p=0.52$ against no record and $1.0$ against V-JEPA~2-AC). The informative pattern is directional: without the record all four lateral reaching trials fail, and with it two succeed. How the lateral action axis maps to the mirrored view is specific to this setup, and four of the seven practice photographs (left, right, and the two diagonals) involve lateral displacement.

\begin{figure}[t]\centering
\includegraphics[width=\columnwidth]{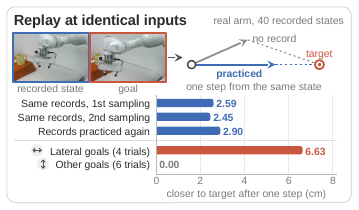}
\caption{Replay at identical inputs on the real arm: how much closer to the target a practiced command ends than a no-record command from the same recorded state. Top: three samplings, the third with records practiced again from scratch. Bottom: the mean of the three, split by goal direction.}
\label{fig:replay}
\end{figure}

\para{Replay at identical inputs.} Since different actions lead to different states, we replay the 40 recorded states of the ten no-record reaching trials, re-sending each original image, end-effector state, and goal image with and without a practice record, and score each returned action by the distance to the target after executing it. This computed distance agrees with the arm: over the 80 recorded steps of both reaching conditions it differs from the next measured position by 0.01\,cm on average, and agrees on whether the step moved closer or farther in 79 of 80 steps. To keep practice photographs away from the scored targets, we split the trials into two halves and practice a separate record for each, from photographs more than 7\,cm from every target that record is scored on. Without the record, the mean distance after a step is 11.24\,cm, averaged over two samples per state. Practiced commands end 2.59 and 2.45\,cm closer in two samplings of the same records and 2.90\,cm with records practiced again from scratch, 2.65\,cm on average (\cref{fig:replay}, top). On the four lateral-goal trials they end 6.63\,cm closer per step, on the six others 0.00\,cm (\cref{fig:replay}, bottom); six of the ten trials improve. The replay agrees with the executed trials on where the record matters. Three separately practiced sets of robot records (the executed-trial one and the two replay sets) all improve on no record, and the three replay samplings lie within 0.45\,cm of each other.

\section{Discussion}
\label{sec:discussion}

\para{One record, two uses.} The same preparation protocol supports forecasting and direct control. A pose-rendered probe finds accurate future placements in both DINOv2 and LeWM spaces; a goal-image query instead yields actions that retain performance on new shapes. Prediction quality and control success are distinct evaluations, not interchangeable scores. Their conjunction is the useful result: logged experience prepared through prediction serves a second task without adapting weights or collecting goal-conditioned demonstrations.

\para{Limitations.}
We evaluate one VLM, Claude Opus~5.5. In simulation both practiced records, standard and low-effort, improved on no prefix, but we have not measured how much records practiced the same way vary. On the arm, three independently practiced sets all improved on no record. Rendering in the forecasting probe supplies shape geometry and measures placement rather than native feature generation. The world-model comparison fixes task data, episodes, observation schedule, and success criterion, and compares complete systems. It does not equalize pretraining or inference compute or test more frequent replanning. Each decision is one high-effort model call, suiting slow, quasi-static tasks. The arm study is small, and its replay scores single actions. Directly written robot actions need workspace limits and supervision; ours are capped at 7.5\,cm per axis.

\section{Conclusion}
\label{sec:conclusion}

A fixed record of prediction experience can serve a frozen VLM after the question changes from forecasting actions to choosing them. PREP raises PushObj control success from 12.9\% to 65.4\% and keeps 62.7\% on new shapes, while pose-rendered forecasts remain competitive in two learned feature spaces. Logs-only and calibration controls locate useful local information, and practice produces the strongest evaluated record. The result offers a simple route from recorded interactions to reusable visual control.

\clearpage
\setcounter{page}{1}
\maketitlesupplementary
\appendix

\section{Simulation details}
\label{app:sim}

\para{Training data.} Each training shape contributes 5{,}000 trajectories of 100 steps. Test trajectories are collected separately and do not overlap with training. After the exclusions and difficulty alignment of \cref{sec:pushobj}, 50 problems remain per shape.

\para{World models.} DINO-WM: frozen DINOv2 encoder, 20 training epochs, epoch 17 used. LeWM: encoder and predictor trained jointly from scratch for 62 epochs, epoch 62 used. JEPA-WM: official code and Push-T configuration, frozen DINOv2 encoder, 18{,}000 training and 2{,}000 validation trajectories, the validation trajectories drawn per shape from the training-shape data; at total batch 256 the lowest-validation-loss checkpoint is used. Planning: CEM with 300 samples, 30 elites, 30 iterations, five-merged-step horizon, replanning after the 25 planned actions, at most five plans per episode; the cost compares predicted final features with goal features, and DINO-WM and JEPA-WM add a pusher position and velocity term with weight 1:1. JEPA-WM uses the same cost as DINO-WM.

\para{Statistics.} Wilson 95\% intervals use $z=1.96$. New shapes: no record 9.3--20.5\% (21/150), practice record 54.7--70.0\% (94/150), JEPA-WM batch 256 22.0--36.4\% (43/150). Training shapes: practice record 60.7--73.6\% (135/200), JEPA-WM batch 256 56.1--69.4\% (126/200). Two-sided Fisher exact tests: practice record versus no record over all episodes, $p<10^{-40}$; versus JEPA-WM batch 256 on new shapes $p<10^{-8}$, on training shapes $p=0.40$, over all episodes $p<10^{-5}$. Calibration study: record versus calibration $p=0.025$, calibration versus none $p=0.064$, record versus none $p<10^{-4}$, T only $p=0.020$. Effort study (30 small-T/T episodes): standard versus low-effort record $p=0.009$, low-effort record versus none $p=0.25$, standard versus none $p<10^{-4}$. These Fisher tests treat conditions as independent samples.

\section{Forecast probe details}
\label{app:probe}

The 35 forecast problems of \cref{sec:forecast} take five per shape from the 350 test problems. Each gives the frames of the current and two preceding merged steps, the ten executed actions, and the next 25 actions, and asks for the five merged-step boundaries; the model answers from the standard record (practiced at maximal effort) at high effort. The keypoint metric averages Euclidean pixel error over the pusher center and four block points on $224\times224$ images. For feature scoring, each pose answer is rendered in the simulator and encoded by DINOv2 or by LeWM's encoder; each of the three answers per problem is scored separately and the three errors are averaged. The decoded-area analysis of \cref{sec:probe} decodes both DINO-WM's predicted features and the encoded true frames and counts pixels within RGB Euclidean distance 40 of the block color (channels 0--255), separately for merged steps with and without contact.

\section{Content studies}
\label{app:content}

All studies in \cref{sec:prefix} use the control queries of Sec.~3.2 and change only the prefix. The unrelated-context, logs-only, and low-effort comparisons in \cref{fig:matched} (top) use the first 20 test episodes of each new shape (L, T, and +), 60 in all; their no-prefix and standard-record outcomes are the corresponding episodes of the main evaluation, and the unrelated conversation is task-unrelated and matched to the standard record's length. The separate effort study uses 30 episodes of small T and T; the low-effort record has the same four problems and the same true outcomes as the standard record, practiced at low reasoning effort, and the standard record and no prefix are evaluated on the same 30 episodes. The calibration study uses ten episodes each of I, square, and T, with the standard record and no prefix evaluated on the same 30 episodes. The calibration text is written from quantities readable in practice feedback and contains no problem's frames, actions, or outcomes, and no padding.

\section{Real-arm details}
\label{app:arm}

\para{Setup and practice.} The arm is a Franka Emika Panda with its standard gripper. The camera is a single uncalibrated RGB camera at the front right; mirroring and the sign change of lateral components are applied identically for both controller families. Practice photographs (\cref{sec:robotrecord}) are taken in the test scene and camera pose: start, 15\,cm forward, backward, left, and right, and 10\,cm forward-right and backward-left. Two problems each order six of them, with five actions defined by measured end-effector differences. Answers predict end-effector positions and eight keypoints; feedback gives true values, errors, and the true photographs with predicted and true points joined.

\para{Replay.} The 40 recorded states of the ten no-record reaching trials (four steps each) are replayed (\cref{sec:robot}). The distance computation is validated on the 80 recorded steps of both reaching conditions (mean discrepancy 0.01\,cm; 79 of 80 directions agree). Each half of the trials is scored with a record whose practice photographs are more than 7\,cm from all of its targets. The no-record reference averages two samples per state (11.24\,cm). Practiced samplings: two evaluations of the same records and one with records practiced again, giving reductions of 2.59, 2.45, and 2.90\,cm (mean 2.65); lateral goals 6.63\,cm, other goals 0.00\,cm; six of ten trials closer.

\section{Additional results and clarifications}
\label{app:extra}

\subsection{A second JEPA-WM training run}
\label{app:jepab64}

Besides the batch-256 run of \cref{tab:main}, JEPA-WM was trained a second time at total batch size 64, with the same 18{,}000/2{,}000 split, the official Push-T settings, and the same checkpoint rule (lowest validation loss). Under the same planner and the same 350 episodes it succeeds on 111 of 200 training-shape episodes (55.5\%; I 27, square 28, Z 30, small T 26) and 31 of 150 new-shape episodes (20.7\%; L 10, T 8, + 13), 142 of 350 overall (40.6\%). Batch 256 is the stronger run on both splits and is the one reported in the main paper. Both runs fall to less than half of their training-shape success on the new shapes (55.5\% to 20.7\%, and 63.0\% to 28.7\%), whereas PREP goes from 67.5\% to 62.7\%.

\subsection{Paired comparisons in the forecast probe}
\label{app:probepaired}

Every condition of the forecast probe answers the same 35 problems, so conditions can also be compared problem by problem. For each problem we take the fifth-step keypoint error averaged over its three answers and apply a two-sided Wilcoxon signed-rank test to the paired differences; the tests were computed post hoc over the pairs of conditions. With high-effort testing fixed, maximal rather than high-effort preparation lowers the error by 7.29 pixels on average when the model writes keypoints (lower on 34 of 35 problems, $p=2.6\times10^{-6}$) and by 9.01 pixels when it writes poses (lower on 31 of 35, $p=9.7\times10^{-7}$). From the maximal-effort record, writing keypoints rather than poses lowers the error by 2.34 pixels (31 of 35 problems, $p=4.5\times10^{-5}$).

\subsection{Computation per decision}
\label{app:cost}

Each control decision is one model call (\cref{sec:method}), and an episode makes at most five. For the world models, each plan runs CEM with 300 sampled action sequences for 30 iterations, 9{,}000 sequences rolled out over five merged steps per decision (\cref{sec:wm}). For V-JEPA~2-AC, its original report gives 16 seconds per action on an RTX 4090~\cite{assran2025}.

\subsection{What the world-model comparison holds fixed}
\label{app:fixed}

\Cref{tab:fixed} lists what is shared by all PushObj controllers and what differs. The differences run in both directions: the VLM brings web-scale pretraining and receives measured block and pusher states in its practice feedback, whereas the world models receive far more task-specific data, pusher state at every decision (DINO-WM and JEPA-WM), and thousands of evaluated candidate sequences per decision. The comparison therefore concerns complete systems under a common task and observation schedule (\cref{sec:discussion}).

\begin{table}[t]
\centering\footnotesize\setlength{\tabcolsep}{2.5pt}
\begin{tabular}{@{}>{\raggedright\arraybackslash}p{0.22\columnwidth}>{\raggedright\arraybackslash}p{0.40\columnwidth}>{\raggedright\arraybackslash}p{0.34\columnwidth}@{}}
\toprule
 & World models & VLM + PREP\\
\midrule
\multicolumn{3}{@{}l}{\emph{Shared}: 350 episodes, starts and goals, success criterion,}\\
\multicolumn{3}{@{}l}{one observation per 25 actions, at most five; no new-shape data.}\\
\midrule
Task data & 20{,}000 trajectories of 100 steps & 4 segments of 35 steps\\
Pretraining & DINOv2 (DINO-WM, JEPA-WM); none (LeWM) & web-scale VLM\\
Labels in preparation & none (no block pose) & measured block and pusher states in feedback\\
Input per decision & 3 frames, actions, goal; + pusher state (DINO-WM, JEPA-WM) & 3 frames, actions, goal\\
Decision & CEM, $300\times30$ sequences & one model call\\
\bottomrule
\end{tabular}
\caption{What the PushObj comparison holds fixed and what differs.}
\label{tab:fixed}
\end{table}

\subsection{Per-shape counts of the prefix studies}
\label{app:prefixcounts}

\Cref{tab:prefixcounts} gives the per-shape successes behind \cref{fig:matched}. In each block, all rows are evaluated on the same episodes.

\begin{table}[t]
\centering\small\setlength{\tabcolsep}{4pt}
\begin{tabular}{@{}lcccc@{}}
\toprule
New-shape set & T & + & L & Total (60)\\
\midrule
No prefix & 5 & 4 & 2 & 11\\
Unrelated, length-matched & 2 & 3 & 1 & 6\\
Logs only & 11 & 10 & 7 & 28\\
Low-effort practice & 9 & 9 & 9 & 27\\
PREP (standard record) & 14 & 14 & 10 & 38\\
\midrule
Calibration set & I & Sq. & T & Total (30)\\
\midrule
No prefix & 3 & 1 & 4 & 8\\
Written calibration & 8 & 5 & 3 & 16\\
PREP (standard record) & 9 & 7 & 9 & 25\\
\bottomrule
\end{tabular}
\caption{Successes per shape in the prefix studies, 20 episodes per shape in the new-shape set and 10 in the calibration set.}
\label{tab:prefixcounts}
\end{table}

\end{document}